\documentclass[fleqn,10pt]{wlscirep}
\usepackage[utf8]{inputenc}
\usepackage[T1]{fontenc}

\usepackage{amsmath}
\usepackage{cmap}
        \usepackage{setspace}
\usepackage{hyperref}
\usepackage[T1]{fontenc}
\usepackage{algorithm}
\usepackage{algpseudocode}
\usepackage{xcolor}
\usepackage{fancyhdr}
\usepackage{comment}
\usepackage{siunitx}

\usepackage{booktabs}
\usepackage{siunitx}

\usepackage[switch]{lineno}

\usepackage{cmap}

\usepackage[T1]{fontenc}
\usepackage{algorithm}
\usepackage{lineno}
\usepackage{amsmath}

\usepackage{lastpage,fancyhdr,graphicx}
\usepackage{epstopdf}

\usepackage{algpseudocode}
\usepackage{xcolor}
\usepackage{fancyhdr}
\usepackage{comment}
\usepackage{enumitem}
\usepackage{float}

\usepackage{graphicx}
\usepackage{pdflscape}
\usepackage{multirow}
\usepackage{lineno}
\usepackage{breakurl}

\usepackage{xspace}
\usepackage{enumitem}
\usepackage{float}
\usepackage{graphicx}
\usepackage{subfigure}
\usepackage{tabularx}
\usepackage{pdflscape}
\usepackage{multirow}
\usepackage{caption}
\usepackage{xspace}

\usepackage{comment}

\newcommand{\hapi}{\texttt{HAPI}\xspace}

\title{Closing the Gap in High-Risk Pregnancy Care Using Machine Learning and Human-AI Collaboration}
\author[1,2,*,+]{Hussein Mozannar}
\author[2,*]{Yuria Utsumi}
\author[3]{Irene Y. Chen}
\author[4]{Stephanie S. Gervasi}
\author[4]{Michele Ewing}
\author[4]{Aaron Smith-McLallen}
\author[2]{David Sontag}

\affil[1]{IDSS, Massachusetts Institute of Technology, Cambridge, MA}
\affil[2]{CSAIL and IMES, Massachusetts Institute of Technology, Cambridge, MA}
\affil[3]{Microsoft Research New England, Cambridge, MA}
\affil[4]{Independence Blue Cross, Philadelphia, Pennsylvania.}
\affil[*]{Equal contribution, co-first authors}
\affil[+]{Corresponding author, \href{mailto:mozannar@mit.edu}{mozannar@mit.edu}}

\begin{abstract}
A high-risk pregnancy is a pregnancy complicated by factors that can adversely affect the outcomes of the mother or the infant. 
Health insurers use algorithms to identify members who would benefit from additional clinical support. This work presents the implementation of a real-world ML-based system to assist care managers in identifying pregnant patients at risk of complications.
In this retrospective evaluation study, we developed a novel hybrid-ML classifier to predict whether patients are pregnant and trained a standard classifier using claims data from a health insurance company in the US to predict whether a patient will develop pregnancy complications.
These models were developed in cooperation with the care management team and integrated into a user interface with explanations for the nurses.
The proposed models outperformed commonly used claim codes for the identification of pregnant patients at the expense of a manageable false positive rate. Our risk complication classifier shows that we can accurately triage patients by risk of complication. Our approach and evaluation are guided by human-centric design. In user studies with the nurses, they preferred the proposed models over existing approaches.
\end{abstract}

\begin{document}
\newgeometry{left=1in, right =1in, top=0.6in, bottom=0.7in}

\maketitle

%
%

\section{Introduction}

\begin{figure*}[!h]
    \centering
        \includegraphics[width=\textwidth]{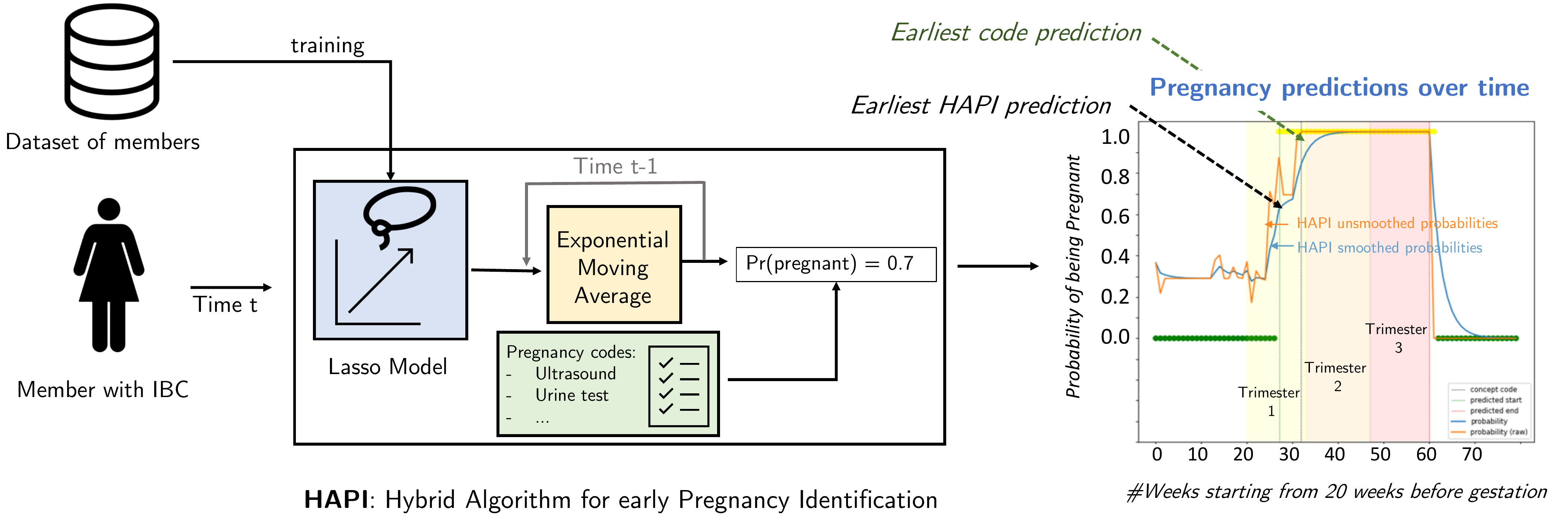}
    \caption{ Illustration of our proposed algorithm \hapi for pregnancy identification. We first collect a historical dataset of members that is used to train the Lasso model that predicts the probability of members being pregnant. Then, at each point in time t in the patient's trajectory (weekly frequency), we pass their claim codes through \hapi which combines the Lasso model and the list of anchor pregnancy codes to obtain a probability of a member being pregnant. We visualize on the rightmost graph  the probability of pregnancy during the member's gestation, where we also show the first instance where there is a code indicating pregnancy start compared to when \hapi predicted pregnancy start.
    }
    \label{fig:patient_level_prediction}
\end{figure*}
High-risk pregnancy is a pregnancy complicated by factors that can adversely affect the health outcomes of the mother, fetus, or infant.
Pregnancy complications like gestational diabetes, hypertension, and pre-eclampsia can lead to childbirth complications such as eclampsia, cardiomyopathy, and embolism and result in adverse pregnancy outcomes, including preterm birth, HELLP syndrome, and intrauterine fetal death.
In 2018, pregnancy and childbirth complications affected 19.6\% and 1.7\% of pregnancies, respectively, in the U.S. \cite{HRPBCBSStats}.
Moreover, systemic disparities in pregnancy and childbirth complications are well-documented.
Black women are significantly more likely to develop preeclampsia and more than three times more likely to die from pregnancy-related complications than White women \cite{petersen_racialethnic_2019}.

Fortunately, timely and appropriate clinical intervention can effectively manage complications during pregnancy and reduce maternal, fetal, and neonatal morbidity and mortality \cite{lassi_essential_2014,teede_gestational_2011,gao_impact_2020,raets_screening_2021,rowan_women_2016}.
Health plan-operated care management programs for high-risk pregnancies aim to coordinate care for at-risk patients across their clinical care team, educate patients about their conditions and medications, and provide education and support managing their conditions \cite{hong_caring_2014,alexander_cost_1999,mate2022field,radin2018healthy}.

\paragraph{Objective.} In this work, we collaborate with the High-Risk Pregnancy (HRP) care management team at an \textbf{A}nonymized Health\textbf{ I}nsurance \textbf{C}ompany (\textbf{AIC}) in the US.  
We aim to improve the member identification process in which nurse case managers review relevant clinical information and make decisions about which members are most appropriate for the HRP program. The process begins with ML algorithms and clinical decision rules to identify pregnant and at-risk members from medical claims, which are served to nurse case managers for review and final determination of program eligibility and appropriateness.
Automated mechanisms for patient risk identification and stratification are critical to efficiently identify pregnant and at-risk patients from a large patient population. We conducted structured interviews with the care managers to understand the identification and stratification process and discover opportunities to improve it. These conversations highlighted that patients being surfaced for evaluation are often no longer pregnant, have a low risk of pregnancy complications, and nurses lack insight into why patients are being surfaced. 

Our first task was to improve the latency with which pregnant patients are identified. Our second task was to accurately identify patients at high risk for pregnancy complications. However, not all complications of pregnancy can be effectively remediated through telephonicly delivered care management. Following the care manager's recommendations, the outreach and education delivered in HRP program would be most impactful for patients with gestational diabetes and gestational hypertension.

\paragraph{Contributions.} This paper presents a recipe for developing automated systems for high-risk pregnancy management programs, from dataset creation to model training and evaluation. We first outline how to build datasets from patient data available to be used to train models for pregnancy identification and detection.
We developed a  novel Hybrid Algorithm for Pregnancy Identification (\hapi) that combines manual code lists with machine learning models. 
 We then train a classifier that predicts the patient's risk for developing complications at each point in their pregnancy.
We integrate these models into a user-friendly interface for nurses to use. We retrospectively evaluate the individual classifiers on over 30k patients, showing we can identify pregnant members earlier on average than predefined code lists and can triage members by risk of complication with an AUC of 0.76. User studies with nurses confirm that the new interface is preferred over existing implementations.

More broadly, we believe our work serves as an important demonstration of human-centric design for ML in healthcare and will be a useful guide for future work in the field.

\section{Related Work}
Much of the existing literature on pregnancy identification focuses on retrospective identification of pregnancy episodes \cite{matcho_inferring_2018,blotiere_development_2018,macdonald_identifying_2019,schink_estimating_2020}. Our goal was to identify pregnancy in a near real-time fashion as information about the patient becomes available through medical and pharmacy claims, lab results, authorizations, and admit, discharge, and transfer data. To the best of our knowledge, we believe this is the first work that accomplishes this objective.  Although there is extensive literature on predicting pregnancy complications using machine learning \cite{bertini2022using,espinosa2021data,islam2022machine,machado2015predicting,li2022improving},  we focus specifically on gestational hypertension and diabetes and making risk predictions as early as possible. While we are aware that certain deep learning architectures perform well for our task, practical considerations limit us to the use of linear classifiers, which perform relatively well. Our approach is to build separate machine learning models for pregnancy start and end identification and risk of pregnancy complications.  When deploying machine learning models in the clinical setting, it is important to provide a rationale for predictions to gain clinicians’ trust and help them make informed decisions \cite{48430,asan2020artificial,reverberi2022experimental,gaube2021ai}.
We discuss other relevant prior work in the remaining sections.

\section{Methods}

\subsection{Dataset Creation For  Pregnancy  Start and End  Identification}

Our approach for identifying the start and end of a patient's pregnancy is based on a machine learning predictor. Since there is no publicly available well-suited data for this task, we built our own dataset to train the model from AIC's members only.
We construct a cohort of female patients with ages between 18 and 48  who had pregnancies with and without complications between 2004 to 2021 but eventually had a live birth. 
We also construct a matching cohort of never-pregnant female patients according to the age distribution of the pregnant sub-cohort.  

To identify pregnant patients for use in our machine learning algorithm to identify pregnancy starts, we use a modified version of the algorithm of Matcho et al. \cite{matcho_inferring_2018} to identify pregnant patients and only select patients who had a healthy live birth.
The original algorithm retrospectively infers the start and end of the most recent pregnancy episode and the corresponding pregnancy outcome or complication. In contrast, our approach identifies gestational episodes in real time.  
We select patients with a live birth only because that allows us to reliably identify the pregnancy start date.
For pregnancies with a live birth without complications, we can reliably identify the pregnancy start date, which we set to be 40 weeks before the end date of pregnancy. For pregnancies with complications, we set the start date to be the first date of occurrence of a pregnancy start code.
The overall dataset consisted of 36735 patients with an average age of 32.3 years composed into three subgroups: 22.6\% pregnancies without complications, 62.4\% pregnancies with complications, and 15.0\% never pregnant.

 For pregnant patients, we extract weekly data starting from  
20 weeks before pregnancy starts to 20 weeks after the pregnancy ends: 80 weeks total - 80 total data points per patient. This allows for early pregnancy and non-pregnancy indicators to be learned while avoiding signals from previous pregnancies. For never-pregnant patients, we sample 80 weeks of data, around the midpoint of their medical history. 
For each data point, we generate non-temporal and temporal features from medical data. For temporal data, we construct windowed features, which aggregate the data within a specified backward time window and map them to a binary indicator feature indicating whether the billing codes occurred or not during that time window. Windowed features for 5-day and 10-day windows are generated using \verb!omop-learn! \cite{kodialam2021deep} for the following categories: medical conditions, prescriptions, procedures, specialty visits, and labs.  We also include 12 non-temporal features, which include age, race, and gender. This gives us a feature set of 62,734 features.
For each subgroup, we split the data into a train set (50\%), validation set (25\%), and test set (25\%) by patients, so no patient data is shared across the different splits. We aggregate all three sub-cohorts to construct the train, validation, and test splits.
A summary of the dataset is provided in Table \ref{tab:table1}. Further details about the dataset creation are found in the Appendix.

\begin{table*}[!ht]
    \centering
        \caption{Summary of patient characteristics and feature processing for the dataset used to build models to identify if a patient is pregnant (Identification Dataset) and for the dataset used to triage patients by risk of complication (Complications Dataset).}
    \begin{tabular}{p{0.3\textwidth}p{0.3\textwidth}p{0.3\textwidth}}
    \hline
        \textbf{Characteristics} & \textbf{Identification Dataset} & \textbf{Complications Dataset}  \\ \hline
         No. of patients & 36,735 & 12,243  \\ \hline
         Race / Ethnicity (\%) & 39.1\% White, 5.7\% Black, 3.4\% Other (rest is unreported)  & 43.8\% White, 5.70\% Black and 3.6\% Other (rest is unreported) \\ \hline
         Average Age in years   & 32.3 ($\sigma$=6.1) & 32.0 ($\sigma$=6.1) \\ \hline
         Pregnancy Complication \%  & 22.6\% without complication, 62.4\% with complications, 15.0\% not pregnant  & 73.6\% without complication, 26.4\% with complications divided into 16.9\% with gestational hypertension and 9.4\% gestational diabetes \\ \hline 
         Dataset split & 50\% training, 25\% validation and 25\% testing & 60\% training, 20\% validation and 20\% testing \\ \hline
         Features generated & $\{5,10\}$ day windowed features and 12 non-temporal features & $\{30,180,365,730,10k\}$ day windowed features and 12 non-temporal features \\ \hline
         Total number of features per patient data point & 112,322  & 62,734 \\ \hline 
    \end{tabular}

    \label{tab:table1}
\end{table*}

\subsection{Algorithm For Pregnancy  Start and End  Identification}

We propose a Hybrid Algorithm for Pregnancy Identification (\hapi) that predicts at each week the probability that a patient is pregnant. The \hapi algorithm predicts a score in $[0,1]$ of the likelihood of the patient being pregnant at each point in time $t$ using their features up to time $t$: $X_t$. 
 \texttt{HAPI} first relies on a set of carefully chosen clinical codes that indicate either the start or end of pregnancy denoted as 'anchors'. Starting from each week of the patient's data, if a code indicating the start of pregnancy is available, we set the start of pregnancy at the first week when the code is available, similarly for codes indicating the end of pregnancy. Otherwise, we use a Lasso regularized logistic regression model \cite{tibshirani1996regression} that is trained with the objective of predicting whether the patient is currently pregnant from the features in the dataset. 
Importantly, we use the Anchor\&Learn approach \cite{halpern2016electronic}, where we remove the anchors from the feature set of the Lasso algorithm so that it focuses on signals not captured by the anchors.
After we get the predictions of the Lasso model at time $t$ as $f(X_t) \in [0,1]$, we pass those predictions to an exponential moving average filter to smooth the predictions over time and obtain $\Tilde{f}(X_t)$. We then binarize the predictions using a learned threshold to obtain $\hat{q}(X_t) \in \{0,1\}$. We predict that the patient is pregnant at time $t$ if $\hat{q}(X_t)=1$ and we have two consecutive increases in $\Tilde{f}(X_t)$ (similarly for end-of-pregnancy prediction).
The Lasso model is learned on the training set of the dataset previously described with hyperparameters chosen on the validation set.  A formal description of the algorithm can be found in the Appendix.

\subsection{Dataset Creation for Pregnancy Complication Prediction }

 After pregnant patients are identified, we have to distinguish between those with a high and low likelihood of developing complications. The case management team identified gestational diabetes (GDB) and gestational hypertension (GHT) as specific complications that could be effectively managed within the HRP program. Our approach is to build a calibrated machine learning classifier that given a patient's data can predict the risk of them developing either gestational diabetes or gestational hypertension. Moreover, the classifier can provide a list of the patient features that led to the prediction as a form of explanation. Since there exists no good public data for evaluating and training the classifier, we construct our own dataset from AIC members. 

We first constructed a cohort of pregnant patients using the algorithm of Matcho et al. \cite{matcho_inferring_2018}, we
then collaborated with the nurse case managers to compile a list of codes that indicate pregnancy episodes with gestational diabetes and gestational hypertension. 
We validated the code set with the care management nurses, who hand-labeled outcome codes for a subset of 20 patients, given data up to the end of the pregnancy episode. This allowed us to (1) validate that the existing codes are indicative of the corresponding outcome, and (2) find new codes indicative of an outcome. This enables us to find patients with outcomes of GHT and GDB during their pregnancy.

 We select the subset of patients with outcomes: live birth (no complication), gestational hypertension, and gestational diabetes. We then select a set of 12,243 patients where 73.6\% are live births, 16.9 \% have gestational hypertension and 9.4\% have gestational diabetes.  We divide the patients into 20\% for testing and 80\% for training and validation. For each patient, we construct 10 data points where each data point is a slice of the patient's data up to a cutoff date. We choose the cutoff dates to be uniformly distributed from three months before pregnancy starts until the end of pregnancy, thus covering all three trimesters. Each data point has a label in $\mathcal{Y}=\{0,1,2\}$ indicating respectively no complication, GHT, and GDB. Each data point has temporal and non-temporal features as in the dataset for pregnancy identification. We fit several standard classification algorithms -- Lasso (L1-regularized), ELASTIC-NET (L1 and L2-regularized), and XGBOOST (gradient-boosted tree) by using the training set and the validation set to pick hyperparameters. 
\subsection{Extracting Evidence for Predictions}

It is essential that we provide information as to why a certain patient had the given predictions. We provide a list of claim codes that had the most effect on the model making its prediction and the polarity of the effect. 
For pregnancy identification with \hapi, for each patient, we surface all anchor codes if they are available and we then surface the highest weighted codes (by absolute value) according to the Lasso model.

For predicting the risk of pregnancy complications, the classifier's top codes by model weight include many variants of diabetes and hypertension codes since the prior history of these conditions is highly predictive of GDB and GHT. However, there is a nontrivial number of patients who have no prior history of these conditions, and they may be affected by a different set of risk factors. To better capture these factors, we partition our dataset conditional on prior history of diabetes and hypertension, and train a separate Lasso model on each of the four subsets: no prior history, history of both conditions, and history of either condition alone. We call these models GROUP-Lasso models. For a given patient, the prediction follows the global Lasso model but to extract evidence for the prediction, we extract the highest weighted features from the GROUP-Lasso model that the patient belongs to depending on their prior history of DB or HT.  

\subsection{User Study Design}

Our main evaluation of \hapi and our pregnancy complications classifier is a retrospective evaluation of performance, however, this evaluation does not mirror how the algorithms will be deployed exactly. The algorithms never make the final decision on who is deemed to be pregnant or at risk, rather it is the care managers after reviewing the algorithm's predictions. We design two user studies that simulate how \hapi and the complications classifier will be deployed where a care manager assesses in a simulation environment of patients. All studies in this work were ruled exempt by our IRB.

With input from the nurse care managers, we built a dashboard mimicking the actual dashboard used by the nurses in the HRP program to surface medical history available in insurance claims and other data sources (e.g. visits, diagnosis codes, demographics).  For testing \hapi, we perform a study with a single nurse under two conditions  A) with the predictions and evidence of \hapi and B) without any algorithmic predictions (control).
The nurse makes predictions in each condition for 12 patients. Each trial took up to an hour. 

For testing the pregnancy complications classifier,  we ran three trials where the nurses made decisions on patients
-- A) one without predictions or evidence, B) one with predictions, and C) one with both predictions and evidence, with each of two nurses (referred to as Nurse 1 and Nurse 2) from the pregnancy care management program (six trials total).  
In each trial, the nurse makes predictions on 18 patients retrospectively. A sketch of the interface is shown in Fig\ref{fig:preg_dashboard}.  For each patient, we asked the nurse if they would call the patient and which complication the patient would develop.

\begin{figure}[!h]
    \centering
  \includegraphics[width=0.6\textwidth]{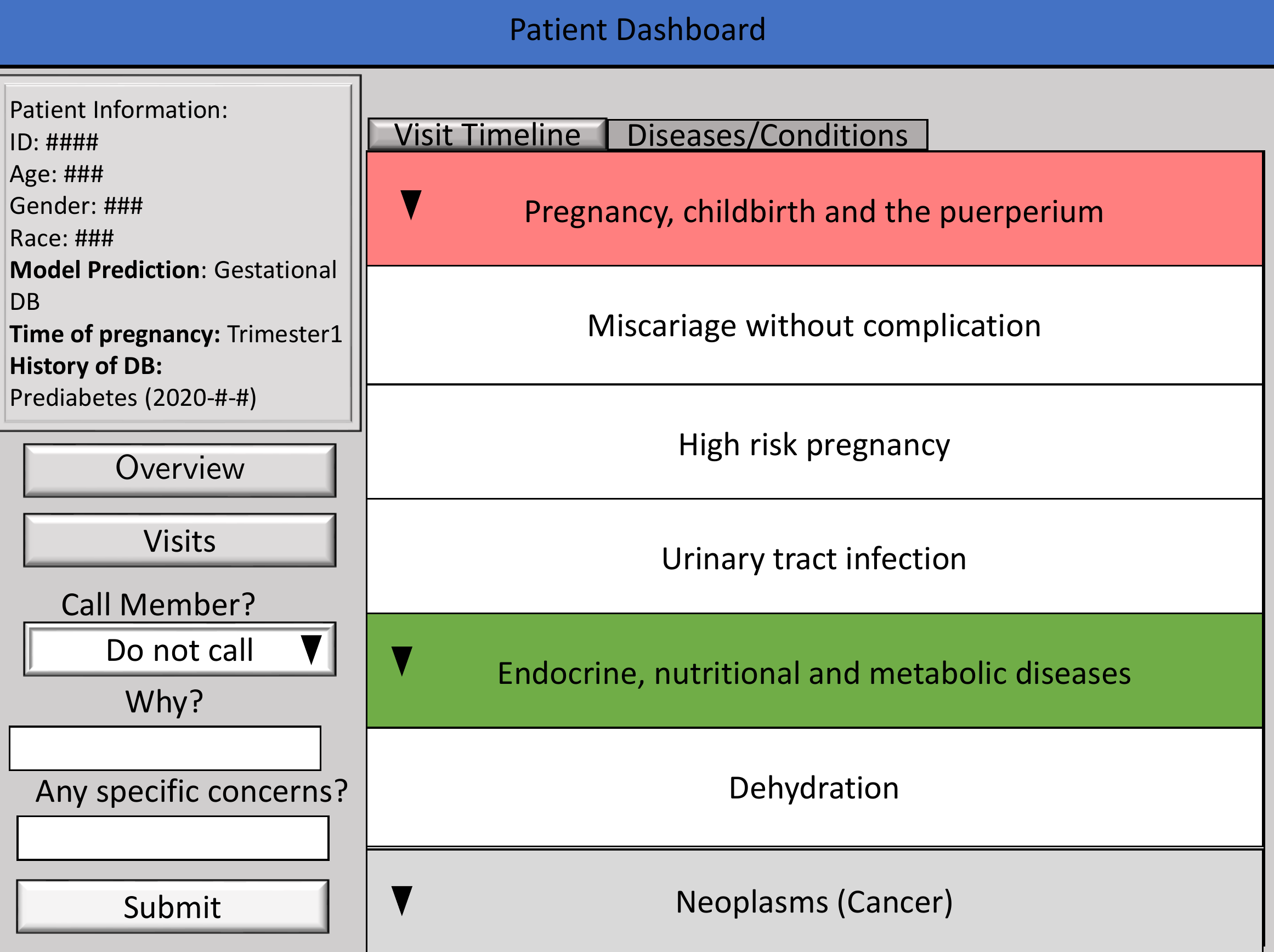} 
    \caption{Patient dashboard sketch for the user study on pregnancy complications classification.  The user interface consists of a left panel containing demographic information and two views: Overview and Visits. We show the subtab Diseases/Conditions from the overview view where the nurse can find the ICD codes for each condition and disease. On the left panel, patient information is shown, the model prediction, and history of prior complications. We color ICD codes positively associated with red complications (intensity varies with correlation) and those negatively associated with complications with green. }
    \label{fig:preg_dashboard}
\end{figure} 

\subsection{Statistical analysis}
For obtaining  confidence intervals for AUC we use a method computed using a distribution-independent
method based on error rate and the number of positive and negative samples introduced in \cite{cortes2004confidence}. For obtaining confidence intervals for accuracy and FNR/FPR metrics, we use the Wilscon score interval. We use McNemar’s test to compare ordinal data proportions and paired t-tests to compare numerical data. All analysis was conducted in Python 3.8 and using the statsmodels and scipy packages.

\section{Results}
\subsection{Identifying Pregnancies From Claims Data}

 We evaluate \texttt{HAPI} on a test set of 9183 patients randomly selected from the dataset. We compare the performance of HAPI against the baseline of only using anchor pregnancy codes for the detection of pregnancy start \cite{macdonald_identifying_2019}. We measure for HAPI and the anchor code list the difference between the predicted start date and the actual pregnancy start date. The actual pregnancy start date is obtained by subtracting 40 weeks from the exact date of birth.

\begin{figure}[h]

  {%
    \subfigure[On all the test set.]{%
      \includegraphics[width=0.45\linewidth]{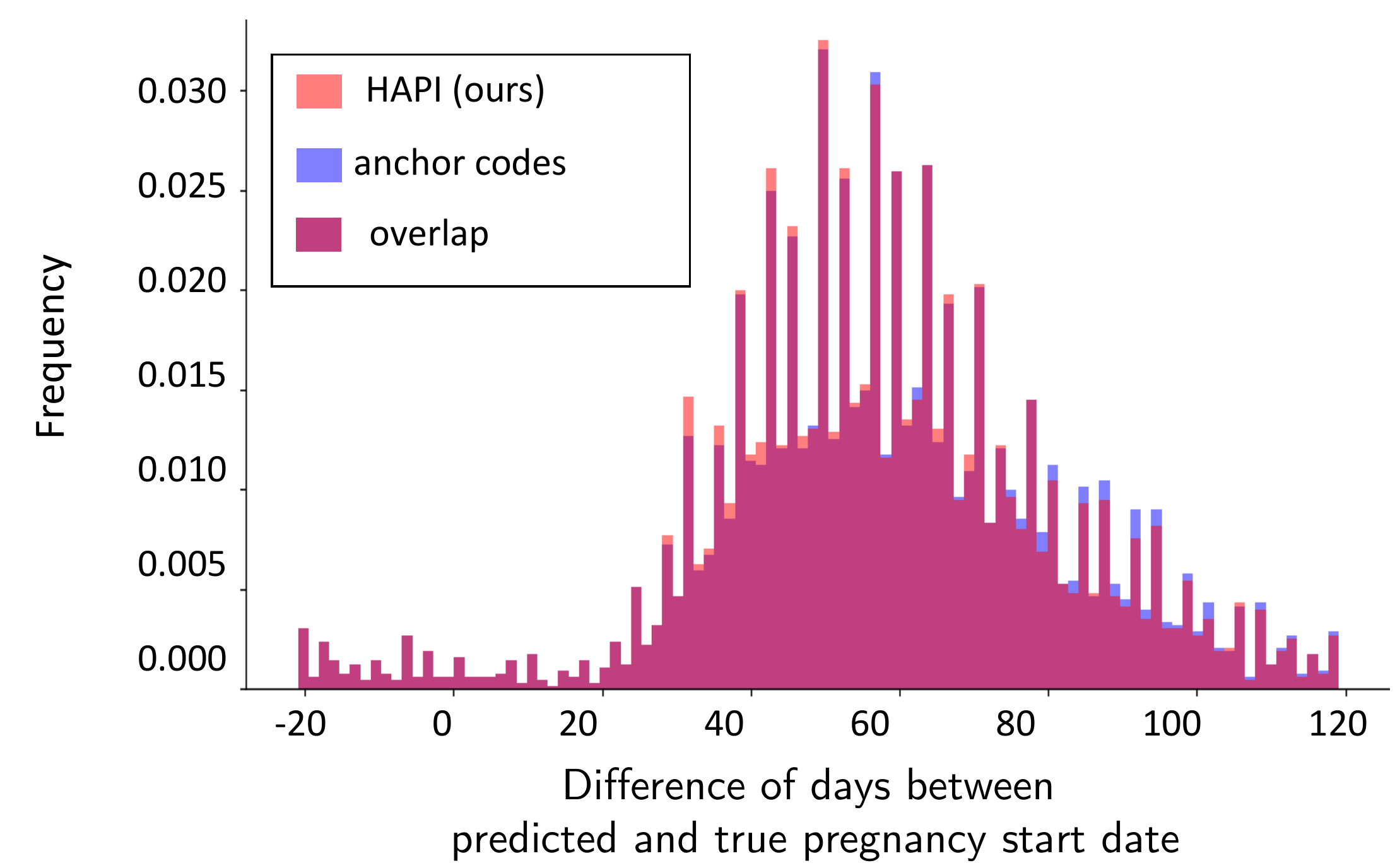}\label{fig:all_test_set}}%
    \qquad 
    \subfigure[On subset of data where \hapi outperforms anchor codes]{%
      \includegraphics[width=0.45\linewidth]{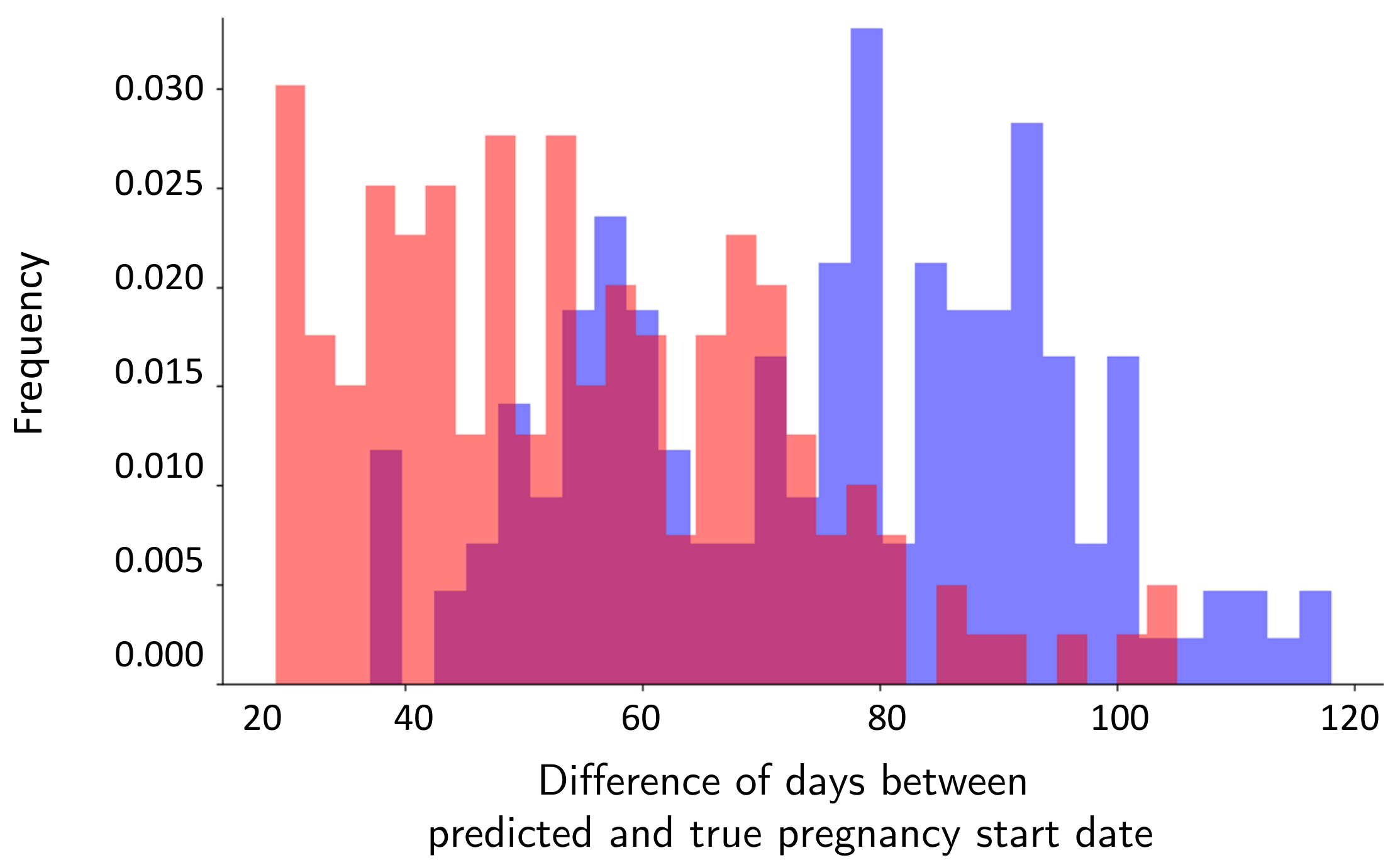}\label{fig:better_subset}}%
  }

    \caption{
   Histogram of pregnancy identification delays for pregnancies with complications for \hapi compared to the anchor codes. We measure the difference of days between the predicted start date and actual start date for our model \hapi compared to a set of predefined pregnancy start codes (anchor codes). In subfigure (a) we show the histogram of differences in all the test patients and we can see that the two distributions overlap. However, in subfigure (b) when we look at the subset of the test patients where \hapi is earlier than the anchor codes ( 3.54\% of the set) we see that \hapi is earlier than the anchor codes.}
    \label{fig:histogram_of_differences_preg_id}
\end{figure}

We show the histogram of the difference between the predicted start date and the actual date for patients with complications in Figure~\ref{fig:histogram_of_differences_preg_id} for both the anchor code list and our proposed algorithm.
Compared to using the code list alone, HAPI predicts an earlier start date for 3.54\% (95\% CI 3.05-4.00, z=14.5, p<0.001)  of patients with pregnancy complications and 4.29\% (95\% CI 3.42-5.16, z=9.6 p<0.001)  earlier for pregnancies without complications. For the patients with complications who are predicted earlier by \hapi, the average difference between the predictions and the actual start date is 54.3 days compared to 75.6 days for the code list ($t=-10\cdot5,p<0\cdot0001$). 
For patients without complications, the average difference is 66.9 days compared to 102.5 days ($t=-6\cdot5,p<0\cdot0001$), respectively. However, when we look at all the test set the average difference is 1 day earlier for  \hapi compared to the code list on patients with and without complications which is not statistically significant.
The model predicts that 5.58\% (95\% CI 4.05-6.40) of non-pregnant patients are in fact pregnant (false positive rate). \texttt{HAPI} can be adjusted using the Lasso model threshold to reduce the false positive rate at the expense of detecting pregnancies later in time.

\subsection{Predicting Pregnancy Complications}

 After pregnant patients are identified, we have to distinguish between those with a high and low likelihood of developing complications.  In Table \ref{tab:risk_eval_populationlvl}, we compare the performance of different machine learning classifiers at predicting whether a patient will develop gestational diabetes or gestational hypertension or neither. We find that the best-performing model in terms of accuracy is a Lasso regularized logistic regression model which achieves an average accuracy of 76.8\% (95\% CI 76.2-77.3) at predicting complications across each test patient pregnancy and  AUC of 0.761 (95\% CI 0.754-767). The Lasso model is able to achieve an accuracy of 73.1\% (95\% CI 72.9-74.2) and AUC of 0.722 (95\% CI 0.710-0.734) when predicting three months before the start of the patient's pregnancy. This indicates that there is a signal at the start of the pregnancy to triage patients by risk of complications.
We assess model performance using data of the patients at different stages of pregnancy and find that accuracy and AUC generally increases as we progress to later pregnancy terms. This indicates that the model performs better as we see more data on the patient, but the confidence intervals overlap in some time periods.

\begin{figure}[h]
    \centering
    \includegraphics[width=0.6\linewidth]{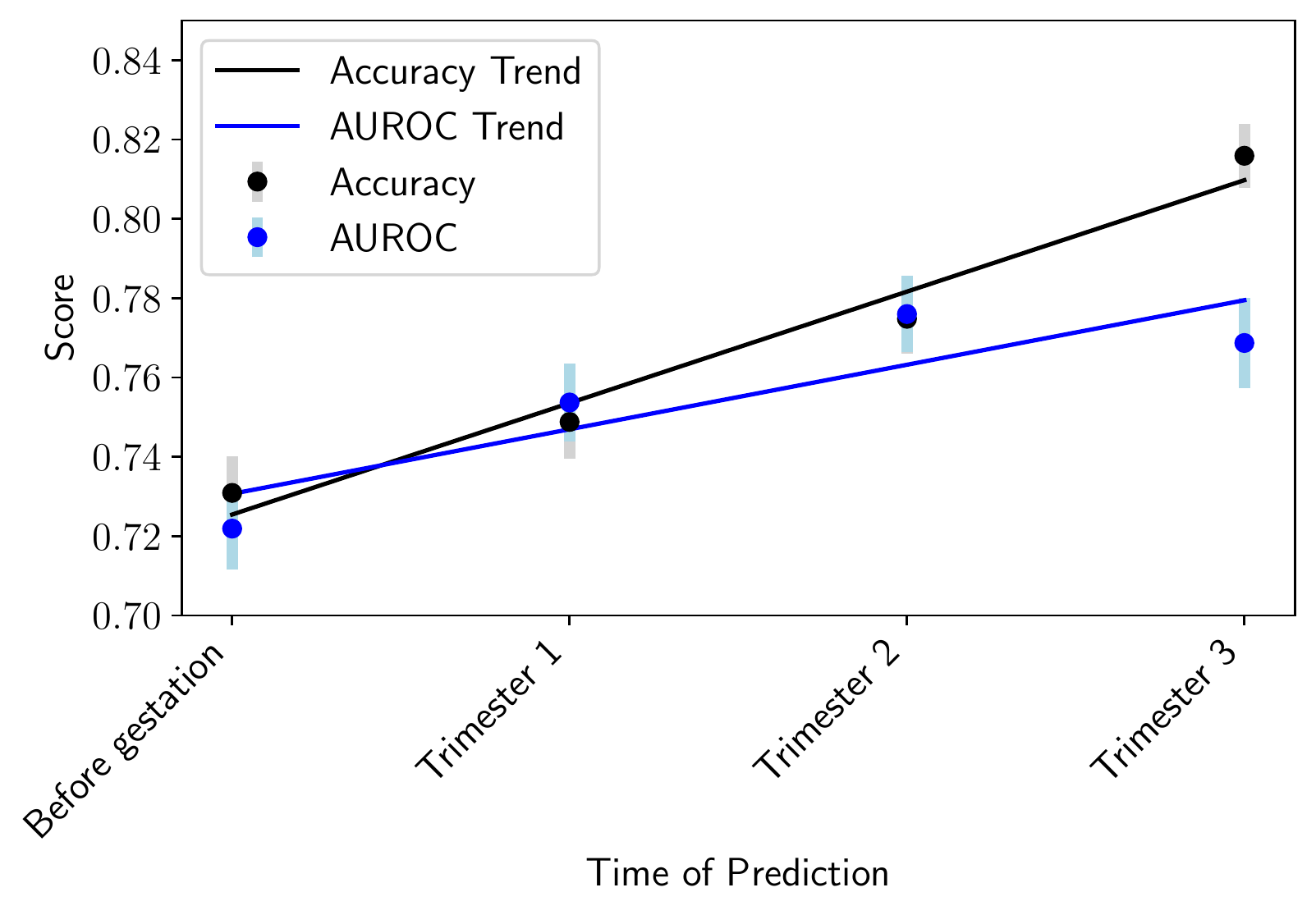}
    \caption{Accuracy and AUROC of the Lasso pregnancy complication predictor as we predict later during pregnancy duration. For each time of prediction, we trim patient data until the time of prediction, we then predict using the trimmed patient data for each time. We plot the linear trend line of the accuracy and AUROC which are shown to be increasing over time, error bars represent 95\% CI. }
    \label{fig:prediction_over_time}
\end{figure}

To assess model performance at different stages of pregnancy, we evaluate the model when predicting on patient's data in each trimester and before gestation. To do this, we trim each member's data until the desired date of prediction and then predict using the Lasso model, results are in Figure~\ref{fig:prediction_over_time}. While confidence intervals do not overlap consistently across time periods, the metrics generally increase as we progress to later pregnancy terms, indicating that the model performs better as we see more data on the member.

 Additionally, we evaluate  how early the model is catching pregnancy complications
While the Lasso model has a high false negative rate of 57.4\% (95\% CI 53.5-61.2), of the patients with true positive predictions (37.6\%), a majority are caught before gestation (59.6\% with 95\% CI 53.1-65.5 ). This is important since early intervention and treatment are important in reducing gestational diabetes and hypertension risk \cite{teede_gestational_2011,raets_screening_2021,rowan_women_2016}. 

\begin{table}[!ht]
    \centering
        \caption{  Evaluation metrics for predictors of pregnancy complications on the test when predicting across four time periods for each patient: before pregnancy, during trimesters 1,2 and 3; results are aggregated across the four periods. We show three different machine learning models, their accuracy, and their AUC on the test set. We provide 95\% confidence intervals obtained for accuracy using the Wilson score interval and for AUC using the method in \cite{cortes2004confidence}. }
    \begin{tabular}{p{3cm}cp{2cm}cp{2cm}}
        \hline

        & \multicolumn{2}{c|}{\textbf{Accuracy}} &  \multicolumn{2}{c|}{\textbf{AUC}}  \\
        \hline
        & \textbf{Mean} & \textbf{95\% CI} &  \textbf{Mean} & \textbf{95\% CI} \\
        \hline
        Lasso (L1) & 0.768 & 0.762-0.773 & 0.761 & 0.754-0.767 \\
        ELASTIC-NET (L1+L2) & 0.713 & 0.707-0.719 & 0.736 & 0.729-0.742 \\
        XGBOOST & 0.687 & 0.681-0.692 & 0.770 & 0.764-0.775 \\
 
        \hline
    \end{tabular}

    \label{tab:risk_eval_populationlvl}
\end{table}

\subsection{Bias/Fairness Audit For Pregnancy Complication Classifier}

 Prior work has shown that care management risk algorithms may contain racial bias due to nuances in how outcomes are defined \cite{obermeyer2019dissecting}. Moreover, there exist systemic health disparities in maternal and infant mortality rates, e.g. Black people have mortality rates over three times higher than White people during pregnancy (40.8 v. 12.7 per 100,000 live births) \cite{petersen_racialethnic_2019}. To this end, we audit our algorithm for potential racial bias. 
We report evaluation metrics in Table \ref{tab:risk_eval_fairness} for the three most common race groups (White - 43.8\%, Black - 5.7\%, Other  - 3.6\%). Other race category includes race outside of the following: American Indian or Alaska Native, Black or African American, White, Asian, Hispanic or Latino, Native Hawaiian or Other Pacific Islander. We note that accuracy for the White group is 77.4\% (95\% CI 76.7-78.2) compared to a lower accuracy for the Black group at 68.1 (95\% CI 65.6-70.5). However, the AUC for the White group is 0.740 (95\% CI 0.730-75.0) which is lower than that of the Black group  0.787 (95\% CI 0.765-0.808).
This may be due to differences in class distribution, since the Black subgroup has much higher rates of complication (44.0\%), compared to White (24.6\%) and Other (25.9\%) races. True positive rates of catching complications are 36.6\%, 27.1\%, and 30.0\%, for Black, White, and Other subgroups, respectively. 
Race data for this analysis comes from electronic medical records with low coverage for race attribution (only $\sim53\%$ of members have some member-level race attributed to examine bias), so true error rates may differ from those reported here.  he lower accuracy of Black patients compared to White or Other race patients can potentially be explained by different base rates. When different subgroups have different base rates, competing definitions of algorithm fairness may conflict~\cite{chouldechova2018frontiers}. It is important to better understand sources of health disparities, potentially through gathering additional information such as social determinants of health~\cite{mccradden2020ethical}.

\begin{table}[h]
    \centering
        \caption{ Evaluation metrics for the Lasso model on the test set, across different race groups for predicting complications at different points in the pregnancy (averaged from 3 months prior to gestation, trimester 1,2 and 3). Rates of complication in each race group are White - 24.6\%, Black - 44.0\%, Other - 25.9\%. For each race group, we obtain the accuracy and AUC on the subgroup alone with 95\% confidence intervals. }
    \begin{tabular}{cp{3cm}p{3cm}}
        \hline
        & \textbf{Accuracy (95\% CI)} & \textbf{AUC (95\% CI)} \\
        \hline
        White & 0.774 (0.767, 0.782) & 0.740 (0.730, 0.750) \\
        Black & 0.681 (0.656, 0.705) & 0.787 (0.765, 0.808) \\
        Other & 0.792 (0.765, 0.819) & 0.826 (0.798, 0.854) \\

        \hline
    \end{tabular}

    \label{tab:risk_eval_fairness}
\end{table}

\subsection{User Studies}

 In the user study for pregnancy identification, in both conditions, the nurse correctly identified 5 of the 8 pregnant patients. 
 Notably, in each trial, we introduced a patient who was falsely detected by the model to be pregnant, but the care manager was successfully able to recognize this incorrect prediction.

In the user study for pregnancy complications, we note that the inclusion of the model prediction and prior history seemed to improve the nurse's accuracy at predicting whether a patient will develop GDP or GHT. Nurse 1 had an accuracy of 56\% without the model, 72\% with the model prediction only, and 67\% with model prediction and evidence. Similarly, nurse 2 had an accuracy of 33\% in condition A, 56\% in B, and 67\% in C. Note that due to the small sample size of the studies, all increases in accuracy are not statistically significant.
Nurse 1 explained that a prior history of diabetes/hypertension or complications in a previous pregnancy is usually sufficient to make a call, but additional information such as distinct risk factors for complications (e.g. polycystic ovary syndrome) can help them build a better profile of the patient and identify those at risk. Both nurses indicated that highlighted evidence helped with obtaining this information more quickly. The evidence helped them focus on important visits and codes, especially when the visit history was lengthy. Nurse 2 said that although not all evidence was useful or made sense, it is easy to filter out the irrelevant ones, i.e. surfacing useful codes should be prioritized over surfacing a few codes. 
 In follow-up interviews and discussions, the nurses expressed a preference for the dashboard used in the user study compared to their previous systems. They noted that the new interface saves an enormous amount of time as they no longer need to access the claims system to review several years of claims data to decipher whether the patient is even pregnant let alone if they have any potential risk factors.

\section{Discussion}

In this study, we developed a machine learning system for the early detection of pregnancy and the identification of high-risk members. This system is part of a real-world deployment at AIC. We introduced a novel algorithm that identifies whether a member is pregnant from insurance claims data by combining indicators for pregnancy start and end with machine learning predictors. We found that it identifies for 3.54\% members an earlier pregnancy start data compared to concept codes and has only a 5.58\% false positive rate.  The model identifies members who may have started pregnancy visits later in their term since, for example, they tested for pregnancy using at-home tests. This could be a reason to offer cost-free pregnancy tests at local clinics so members are incentivized to get tested formally, and in turn, the insurance company obtains data to identify pregnant members earlier. A large proportion of these members also tend to be high risk, which is exactly who we want to identify early for early intervention and treatment. 
 Leveraging this information, we then identified members at the greatest risk for pregnancy complications so that care managers can provide timely and effective support. Using predictors of gestational diabetes and gestational hypertension, our model achieved an AUROC performance of 0.76.

We followed a human-centered design methodology and showed that it can improve the care management program for high-risk pregnancies at IBC. Because care managers are often faced with limited and fragmented interactions with patients, we conducted extensive discussions and interviews with care managers of the HRP program to identify their current needs and greatest challenges. These insights—combined with insurance claims—can help early detection of pregnancy, accurate identification of impactable high-risk members, and provision of explainable indicators to supplement predictions. We show that when actively engaging critical stakeholders like the care managers, machine learning systems can guide care management to prevent pregnancy complications.

We then set up a mock enrollment dashboard and evaluated these methods across two user studies and found two key findings. First, the pregnancy identification algorithm helps nurses identify pregnancies earlier while correctly filtering out false-positive members. Second, showing the pregnancy complication model's prediction and prior history of chronic conditions improves nurses' performance metrics when deciding who to call. While model explanations adversely affected the nurse's performance in terms of time per member and how early they identify pregnant members in the pregnancy identification study, we observed that explanations improved notes about the member in the pregnancy risk factor study without much difference in nurse's classification performance. The latter study better integrated explanations into the clinical workflow, and nurses appeared to disagree with the explanations less, which emphasizes the importance of the explanation method and how they are presented. Our study demonstrated that close collaboration with care managers can be used to leverage insurance claims to improve the care of pregnant patients.
We hope that our results can serve as a call to action for similar predictive models used to allocate care. In a recent report in the Journal of Biomedical Informatics, researchers advocated for more overlap in human-computer interaction and clinical decision-making tasks to improve precision medicine~\cite{rundo2020recent}. Our work expands on those topics to empower the domain experts and primary users of our system. We found that comprehensive needs-finding interviews with the care managers greatly enhanced our targeted ML system. Not only were we able to focus on the most salient problems facing care managers, but the resulting ML system also has better resource allocation for pregnancy patients. 

Our study opens several areas for future work. As with any machine learning system, continual validation of our models across time is key to ensuring robust and generalizable performance. Predictors of early pregnancy detection and predictors of high-risk pregnancy may change over time due to improvements in health technology and patterns of healthcare utilization. Computational work in transfer learning and robustness can help adapt our models over time with minimal adjustment. Additionally, topics of pregnancy may raise questions about patient privacy. Our model keeps patient data completely private except for the minimal set of relevant care managers; however, advances in patient privacy protection may also be relevant.

\section{Limitations and Conclusion}

 There are limitations to our study that need to be addressed. In Appendix Figure~\ref{fig:how_early_risk}, we stratify the population by when our ML system provided a relevant alert. Unfortunately, 60\% of the alerts are never sounded for patients who have complications. This gap in our model performance is likely due to the sparsity of insurance claims and the delay of visits by the patients, both challenges often faced by models working with healthcare data~\cite{chen2021ethical}. 
We are also concerned with the disparate impact of the ML system on different patient subpopulations, particularly historically vulnerable groups. Health insurers are actively creating best practices for auditing and improving algorithmic bias, with the first step being the measurement of existing bias~\cite{gervasi2022potential}. In Table~\ref{tab:risk_eval_fairness}, we show the performance of the detection algorithm on White, Black, and Other race patients. The lower accuracy of Black patients compared to White or Other race patients can potentially be explained by different base rates. When different subgroups have different base rates, competing definitions of algorithm fairness may conflict~\cite{chouldechova2018frontiers}. It is important to better understand sources of health disparities, potentially through gathering additional information such as social determinants of health~\cite{mccradden2020ethical}.

In conclusion, we have developed novel algorithms for the identification of pregnancy and triage of pregnant members by  risk of complication. These algorithms' development and subsequent evaluations followed a  human-centered design methodology with extensive collaboration with the high-risk pregnancy care managers at AIC. Thus, we demonstrated that the active engagement of key stakeholders like care managers can substantially improve the clinical workflow and quality of care given by care managers for pregnant patients.

\section*{Author Contributions}
Conception and design: H.M., Y.U., D.S., S.G., A.S.

Model Development: H.M., Y.U.

User Study Development: Y.U

Data Collection: H.M., Y.U., M.E.

Data analysis: H.M., Y.U.

Data Interpretation: H.M., Y.U., I.C., D.S., S.G., A.S.

Supervision: D.S.

Manuscript writing: H.M., Y.U., I.C., D.S., S.G., A.S., M.E.

\section*{Acknowledgements} 
H.M., Y.U., I.C. and D.S. were supported by a grant from Independence Blue Cross.

\section*{Competing Interests}
The research was financially supported by a grant from Independence Blue Cross, which also contributed the data for the study. The sponsor collected the data, reviewed the manuscript, and approved the decision to submit the manuscript for publication. H.M., Y.U., I.C. and D.S. were supported by the grant. M.E., S.G., A.S. are employees of Independence Blue Cross.

\section*{Data Availability}
The datasets generated and analyzed during the current study are not publicly available as they contain insurance claims data and demographic data (including age and ethnicity) of members insured by Independence Blue Cross and the data is de-identified but not anonymous.

\bibliography{sample}
\clearpage

\appendix

\section{Supplemental Information}
\subsection{Ethics}
For our human subject experiments described in our results with the care managers, we obtained an exempt evaluation from our IRB.  Our IRB judged that our research activities meet the criteria for exemption as defined by Federal regulation 45 CFR
46 under the following:
\begin{itemize}
    \item 
\textbf{ Exempt Category 3 - Benign Behavioral Intervention}
 Research involving benign behavioral interventions where the study activities are limited to
adults only and disclosure of the subjects' responses outside the research could not reasonably
place the subjects at risk for criminal or civil liability or be damaging to the subjects' financial
standing, employability, educational advancement, or reputation. Research does not involve
deception or participants prospectively agree to the deception. 45 CFR 46.104(d)(3)
\item \textbf{Exempt Category 2 - Educational Testing, Surveys, Interviews or Observation}
 Research involving surveys, interviews, educational tests or observation of public behavior
with adults or children and disclosure of the subjects' responses outside the research could not
reasonably place the subjects at risk for criminal or civil liability or be damaging to the subjects'
financial standing, employability, educational advancement, or reputation. Research activities with
children must be limited to educational tests or observation of public behavior and cannot include
direct intervention by the investigator. 45 CFR 46.104(d)(2)
\end{itemize}

\subsection{Dataset Creation Algorithm for Pregnancy Identification}\label{apx:preg_cohort_alg}

\begin{figure}[H]
    \centering
    \includegraphics[width=\textwidth]{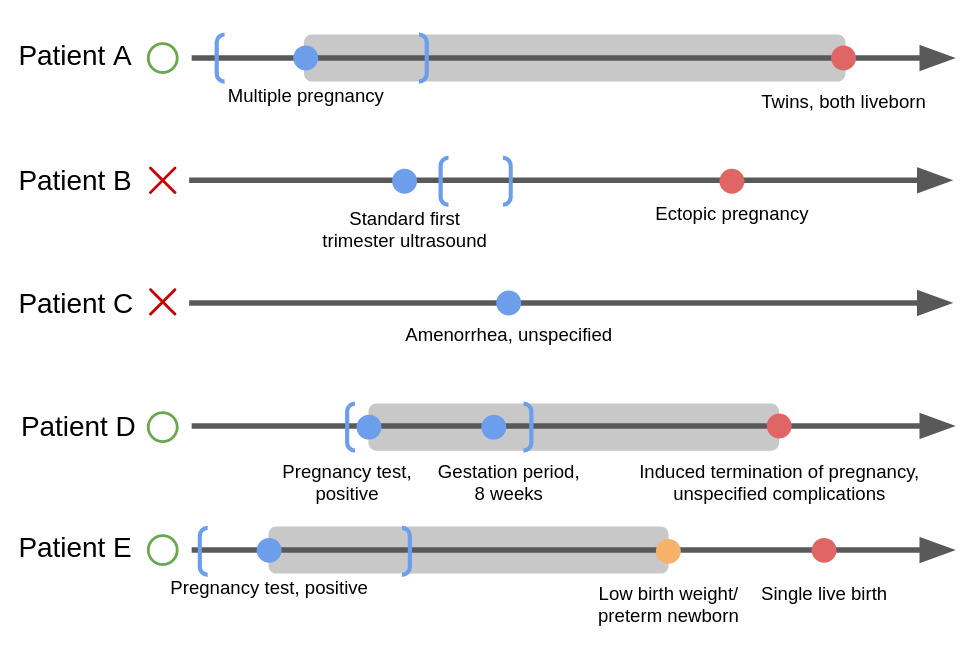}
    \caption{Illustration of the pregnancy cohort selection algorithm (\ref{alg:build_preg_cohort}). First, the most recent pregnancy outcome is detected (red point), referencing outcome codes defined in \cite{matcho_inferring_2018}. Then, we search for pregnancy start code(s) (blue point(s)) within a specified lookback window for the corresponding outcome \cite{matcho_inferring_2018} (blue brackets); the earliest start code marks the start of that pregnancy episode. Finally, we do a forward search for any additional pregnancy outcome or complications, referencing additional outcome codes compiled internally at AIC (orange point); if one exists, the pregnancy outcome is updated. \newline\newline Member B is excluded from the cohort since no pregnancy start code was detected within the lookback window. Member C is excluded since there was no associated pregnancy outcome code; amenorrhea alone cannot indicate pregnancy has started since it can be caused by non-pregnancy-related factors (e.g. stress, menopause).}
    \label{fig:preg_cohort_selection}
\end{figure}

We build on \cite{matcho_inferring_2018}, which presents an algorithm for inferring pregnancy episodes across a set of pregnancy outcomes in OMOP Common Data Model. Our modified algorithm can handle a larger set of pregnancy outcomes, e.g. neonatal ICU admission, by doing a forward search to update the outcome once the pregnancy episode is identified. We describe our modified version in Algorithm \ref{alg:build_preg_cohort}. We illustrate the algorithm in Figure \ref{fig:preg_cohort_selection} and present a subset of target codes for reference in \ref{tab:build_preg_cohort_target_pass2}.

\begin{figure}[H]
    \centering
    \begin{tabular}{lc}
        (a)&\includegraphics[width=\textwidth]{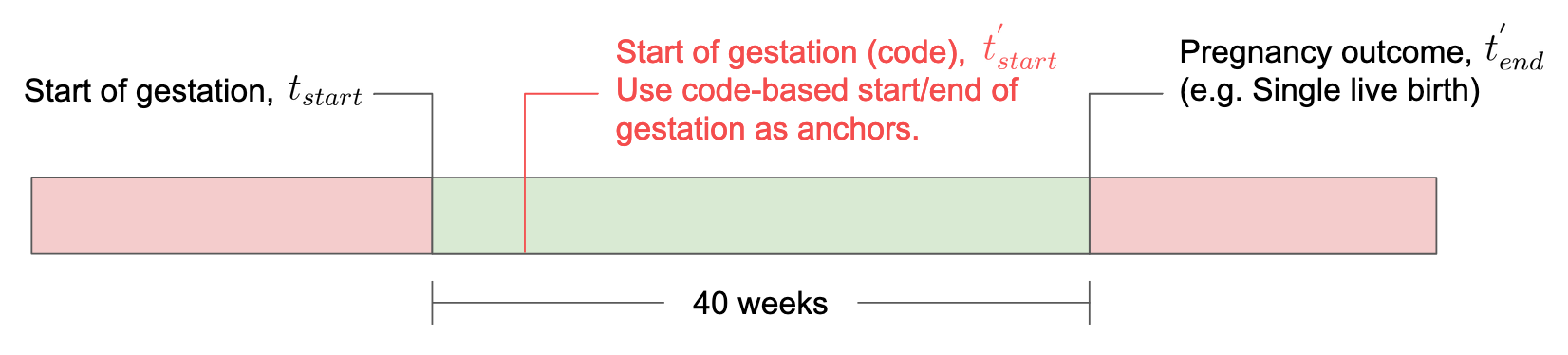}\\
        (b)&\includegraphics[width=\textwidth]{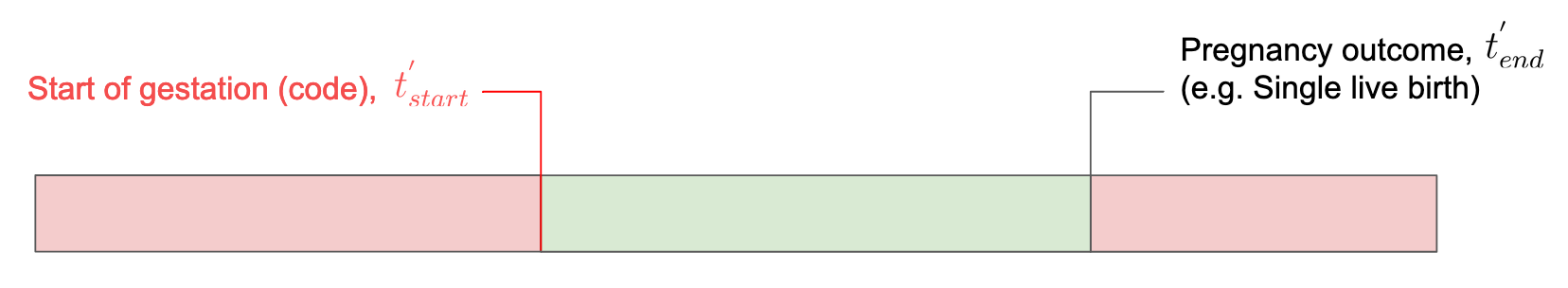}
    \end{tabular}
    \caption{Labeling start and end of each pregnancy episode. (a) For pregnancies without complications, we set the start of gestation to be 40 weeks prior to when the outcome code is observed ($t_{start}=t^{'}_{end}-40\text{ weeks}$), assuming a full term pregnancy. (b) For pregnancies with complications, we set the start of gestation to be the date of pregnancy start code ($t_{start}=t^{'}_{start}$). 
    These points give us a reference frame for data sampling, labeling, and evaluation inside and outside of pregnancy.}
    \label{fig:preg_episode_labeling}
\end{figure}
\clearpage

\begin{table}[H]
    \centering
    \begin{tabular}{|p{3cm}|p{7.5cm}|p{2cm}|p{1.75cm}|}
        \hline
        \textbf{Outcome} & \textbf{Target Codes} & \textbf{Pregnancy ID?} & \textbf{Risk Factors?} \\
        \hline
        {\small Neonatal Intensive Care Unit (NICU)} & {\tiny Newborn light for gestational age\newline Low birth weight infant\newline Birth injury to central nervous system\newline Respiratory distress syndrome in the newborn\newline Pulmonary hypertension of newborn} & X & X \\
        {\small Hypertension/Pre-eclampsia (HPPE)} & {\tiny Pre-existing hypertension in obstetric context\newline Transient hypertension of pregnancy\newline Renal hypertension complicating pregnancy\newline Severe pre-eclampsia\newline Gestational proteinuria} & X & X \\
        {\small Pre-term birth} & {\tiny Preterm premature rupture of membranes\newline Fetal or neonatal effect of maternal premature rupture of membrane\newline Baby premature, 24-26 weeks\newline Extreme immaturity, 750-999 grams\newline Metabolic bone disease of prematurity} & X & X \\
        {\small Gestational Hypertension} & {\tiny Unspecified maternal hypertension\newline Gestational [pregnancy-induced] hypertension\newline Hypertension, Pregnancy-Induced\newline gestational proteinuria\newline Mild to moderate pre-eclampsia} &  & X \\
        {\small Gestational Diabetes} & {\tiny Gestational diabetes mellitus in childbirth\newline Diabetes mellitus arising in pregnancy\newline Gestational diabetes mellitus in the puerperium\newline Gestational diabetes mellitus complicating pregnancy\newline Maternal gestational diabetes mellitus} &  & X \\
        \hline
    \end{tabular}
    \caption{Pregnancy outcomes and examples of corresponding target codes and indicators of whether the outcome was included in the second pass search during cohort creation for pregnancy identification and pregnancy risk factors.}
    \label{tab:build_preg_cohort_target_pass2}
\end{table}

\begin{algorithm}
\caption{Building pregnant cohort. }\label{alg:build_preg_cohort}
\begin{algorithmic}
\For{$i\in P$}
\State \textit{// Detect and classify most recent pregnancy outcome (first pass)}
\State $t_{out}^{i}, outcome^{i}\gets getPregnancyOutcome(\mathcal{H}^{i})$
\State
\State \textit{// Backtrack to estimate pregnancy start}
\State $t_{start, min}\gets t_{out}^{i}-g_{max}^{outcome}$\Comment{Lower bound for pregnancy start}
\State $t_{start, max}\gets t_{out}^{i}-g_{min}^{outcome}$\Comment{Upper bound for pregnancy start}
\State $t_{start}^{i}\gets estimatePregnancyStart(t_{start, min}, t_{start, max})$
\State
\State \textit{// Forward search to update pregnancy outcome (second pass) }
\State $t_{out}^{i}, outcome^{i}\gets updatePregnancyOutcome(t^{i}_{start}, t^{i}_{out})$
\EndFor
\end{algorithmic}
\end{algorithm}

We build a cohort of patients who were never pregnant throughout their claims history. We sample these patients according to the age distribution of pregnant members (mean: 31.8 years, standard deviation: 4.8 years) and define ``never pregnant'' to be any member who does not have any of the pregnancy start or outcome concept codes present in their claims history.

\subsection{Dataset Creation Algorithm for Pregnancy Complication Prediction}\label{apx:dataset_preg_risk}

In Algorithm \ref{alg:build_preg_cohort}, the first pass phase that searches for the most recent pregnancy outcome references the original pregnancy outcomes and corresponding target codes defined in \cite{matcho_inferring_2018}. In the second pass phase performs a second search to update the previous outcome, we reference target codes for additional outcomes. We present a subset of target codes for these outcomes and an indicator for when they are used in Table \ref{tab:build_preg_cohort_target_pass2}

We queried for pregnancy episodes with a gestational diabetes ICD 10 code (O24.11-O24.93) using ATLAS \cite{ohdsi-atlas}.  We then filtered for unique diagnosis codes within those episodes and selected the most frequently occurring diagnosis codes as the initial set of target codes for gestational diabetes outcomes. The same procedure was repeated for gestational hypertension/pre-eclampsia (ICD 10 code O10.011-O16.9). 
We validated the code set with the care management nurses, who hand-labeled outcome codes for a subset of 20 members, given data up to the end of the pregnancy episode. This allowed us to (1) validate that the existing codes are indicative of the corresponding outcome, and (2) find new codes indicative of an outcome. For example, Methyldopa 250 MG Oral Tablet, an anti-hypertensive drug, was added as a code for gestational HT/PE.

Similar to pregnancy identification, we generate non-temporal and temporal features for each sampled point. For temporal data, we generate windowed features for 30 day, 180 day, 365 day, 730 day, and 10k day windows using \verb!omop-learn! for the following categories: medical conditions, prescriptions, procedures, specialty visits, and labs. We also include 12 non-temporal features, which include age, race, and gender. This gives us a feature set of 112,322 features.

\subsection{Hyperparameter Selection for Machine Learning Models}\label{apx:hyerparams}

For the pregnancy identification LASSO model, we report the hyperparameter search space in Table \ref{tab:preg_id_hyperparams}. We select the model with the highest validation accuracy. The decision threshold is chosen to be the geometric mean of sensitivity and specificity on the validation set.

\begin{table}[H]
    \centering
    \begin{tabular}{cc}
        \hline
        \textbf{Hyperparameters} & \textbf{Search Range} \\
        \hline
        Regularization strength (C) & 1e-3, 7.5e-4, 5e-4, 2.5e-4*, 1e-4 \\
        Tolerance & 1*, 1e-1, 1e-2, 1e-3, 1e-4 \\
        \hline
    \end{tabular}
    \caption{Hyperparameter search range for pregnancy identification model. Asterisk marks the chosen hyperparameters.}
    \label{tab:preg_id_hyperparams}
\end{table}

For the pregnancy complications models, we report the hyperparameter search space in Table \ref{tab:preg_risk_hyperparams}. Note that we also correct for class imbalance by weighting each class $j$ by $p(y_j)^{-1}$, where $p(y_j)$ is the proportion of outcomes under class $j$ in the training set. We select the model with the highest product of AUROC and accuracy on the validation set. 

\begin{table}[H]
    \centering
    \begin{tabular}{ccc}
        \hline
         & \textbf{Hyperparameters} & \textbf{Search Range} \\
        \hline
        \multirow{2}{*}{\textbf{LASSO}} & Regularization strength (C) & 1, 1e-1, 1e-2, 1e-3*, 1e-4 \\
        \multirow{2}{*}{} & Tolerance & 1e-1, 1e-2, 1e-3*, 1e-4 \\
        \hline
        \multirow{2}{*}{\textbf{ELASTIC-NET}} & L1-ratio & 0.25*, 0.5, 0.75 \\
        \multirow{2}{*}{} & Tolerance & 1e-1*, 5e-2 \\
        \hline
        \multirow{1}{*}{\textbf{XGBOOST}} & Learning rate & 1e-1*, 1e-2, 1e-3, 1e-4 \\
        \hline
    \end{tabular}
    \caption{Hyperparameter search range for pregnancy risk model. Asterisk marks the chosen hyperparameters.}
    \label{tab:preg_risk_hyperparams}
\end{table}

\subsection{Algorithm For Pregnancy Identification}\label{apx:ag_preg_id}

We formally describe our pregnancy identification algorithm continuing on from the Methods section. 

We combine the anchors and the Lasso model into a hybrid model $f(X_t)$ that does the following: if there exists a pregnancy start code only then $f(X_t)=1$, if there is a pregnancy end code then $f(X_t)=0$, otherwise $f(X_t)$ follows the prediction of the Lasso model. 
After we get the predictions at time $t$ as $f(X_t)$ we pass those predictions to an exponential moving average filter. This serves to smooth the predictions over the last 5 time points with a decay factor of $1/3$ to get a result $\hat{q}$. We then binarize $\hat{q}$ with a learned threshold $\tau$ chosen to maximize the geometric mean of the F1-score of the pregnancy predictions to obtain a binary prediction $\hat{y}$. This process is performed at each time stamp for the member's data. We predict the pregnancy start date as the first instance of time where $\hat{y}$ is $1$ and we have two consecutive increase scores $\hat{q}$, similarly, we predict the pregnancy end date as the instance of time $\hat{y}$ is $0$ and we have two consecutive decreasing scores $\hat{q}$ (given we already predicted the start).

\begin{algorithm}[H]
\caption{Inferring pregnancy start and end for each member. }\label{alg:infer_pregnancy_for_all}
\begin{algorithmic}
\For{$i\in P$}
    \State $\hat{p} \gets f(X_i)$ \Comment{predict probability of pregnancy over time}
    \State $\hat{q} \gets$\verb!EMA!$(\hat{p})$ \Comment{smooth with exponential moving average filter}
    \State $\hat{y} \gets \verb!predict!(\hat{q})$ \Comment{returns binary predictions}
    \State $\hat{start}_i, \hat{end}_i \gets$\verb!InferEpisode!$(\hat{q}, \hat{y})$ \Comment{infer pregnancy start and end (see Alg. \ref{alg:infer_eps})}
\EndFor
\end{algorithmic}
\end{algorithm}

\begin{algorithm}[H]
\caption{Inferring pregnancy start and end, given smoothed probability and predictions over time ($\hat{q},\hat{y}$).}\label{alg:infer_eps}
\begin{algorithmic}
\State isStart=True; $l$=len($\hat{q}$)
\State start, end = None, None
\For{$t=0:l-2$}
    \If {(isStart) and $(\hat{y}[t]==1)$ and $(\hat{q}[t]<\hat{q}[t+1])$} 
        \State \textit{// set pregnancy start if we have +ve prediction and increasing probability}
        \State start$\gets t+1$; isStart$\gets$False
    \ElsIf{(not isStart) and $(\hat{y}[t]==0)$ and $(\hat{q}[t]>\hat{q}[t+1])$)}
        \State \textit{// set pregnancy end if we have -ve prediction and decreasing probability}
        \State end$\gets t+1$
    \EndIf
    \\
    \State \textit{// use code-based prediction by default if pregnancy start is before }
    \State \textit{// 1 month after true pregnancy start (and we are simulating nurses filtering)}
    \If{nurseFilter and start < trueStart+deltaMonth}
    start$\gets$codeStart
    \EndIf
    \\
    \State \textit{// set }\text{start}\textit{ and }\text{end}\textit{ to be the earliest value (code-based or model-based)}
    \State start$\gets$ min(codeStart, start)
    \State end$\gets$ min(codeEnd, end)
\EndFor
\end{algorithmic}
\end{algorithm}

\subsection{Additional Results for Pregnancy Identification Retrospective Evaluation}\label{apx:additional_results_id}

We include additional results for our retrospective evaluation of the pregnancy identification algorithm.

\begin{table}[H]
    \centering
    \small
    \begin{tabular}{|p{\textwidth}|}
        \hline
        \textbf{Feature name} \\
        \hline
        2213418 - procedure - Immunization administration (includes percutaneous, intradermal, subcutaneous, or intramuscular injections); 1 vaccine (single or combination vaccine/toxoid)  \\
        2212167 - labs - Urinalysis, by dip stick or tablet reagent for bilirubin, glucose, hemoglobin, ketones, leukocytes, nitrite, pH, protein, specific gravity, urobilinogen, any number of these constituents; non-automated, without microscopy \\
        2108115 - procedure - Collection of venous blood by venipuncture \\
        3050479 - labs - Immature granulocytes/100 leukocytes in Blood \\
        2212996 - labs - Culture, bacterial; quantitative colony count, urine \\
        3033575 - labs - Monocytes [\#/volume] in Blood by Automated count \\
        3023314 - labs - Hematocrit [Volume Fraction] of Blood by Automated count \\
        3014576 - labs - Chloride [Moles/volume] in Serum or Plasma \\
        38004461 - specialty - Obstetrics/Gynecology \\
        3015746 - labs - Specimen source identified \\
        \hline
    \end{tabular}
    \caption{Top positive features surfaced by non-pregnant members who were inferred to be pregnant.}
    \label{tab:fp_top_features}
\end{table}

\begin{figure}[htbp]
  {\caption{ Histogram of pregnancy identification delays for pregnancies without complications for \hapi compared to the codes. We measure the difference of days between the predicted start date and actual start date for our model \hapi compared to a set of predefined pregnancy start codes (anchor codes). In subfigure (a) we show the histogram of differences on all the test and we can see that the two distributions overlap. However, in subfigure (b) when we look at the subset of the test where \hapi is earlier than the  codes.}}
  {%
    \subfigure[On all the test set.]{%
      \includegraphics[width=0.47\linewidth]{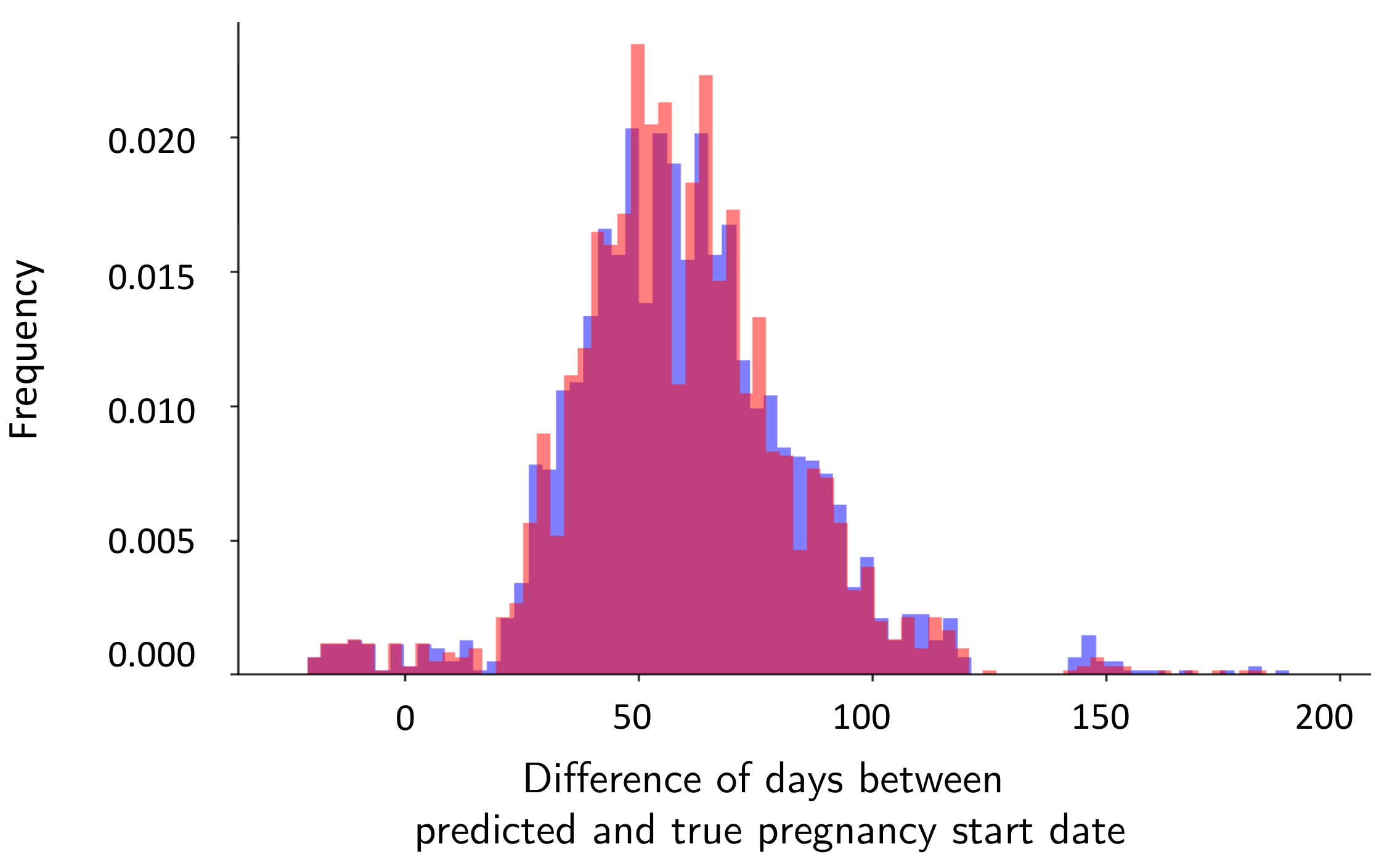}\label{fig:all_test_set}}%
    \qquad 
    \subfigure[On subset of data where \hapi outperforms anchor codes]{%
      \includegraphics[width=0.47\linewidth]{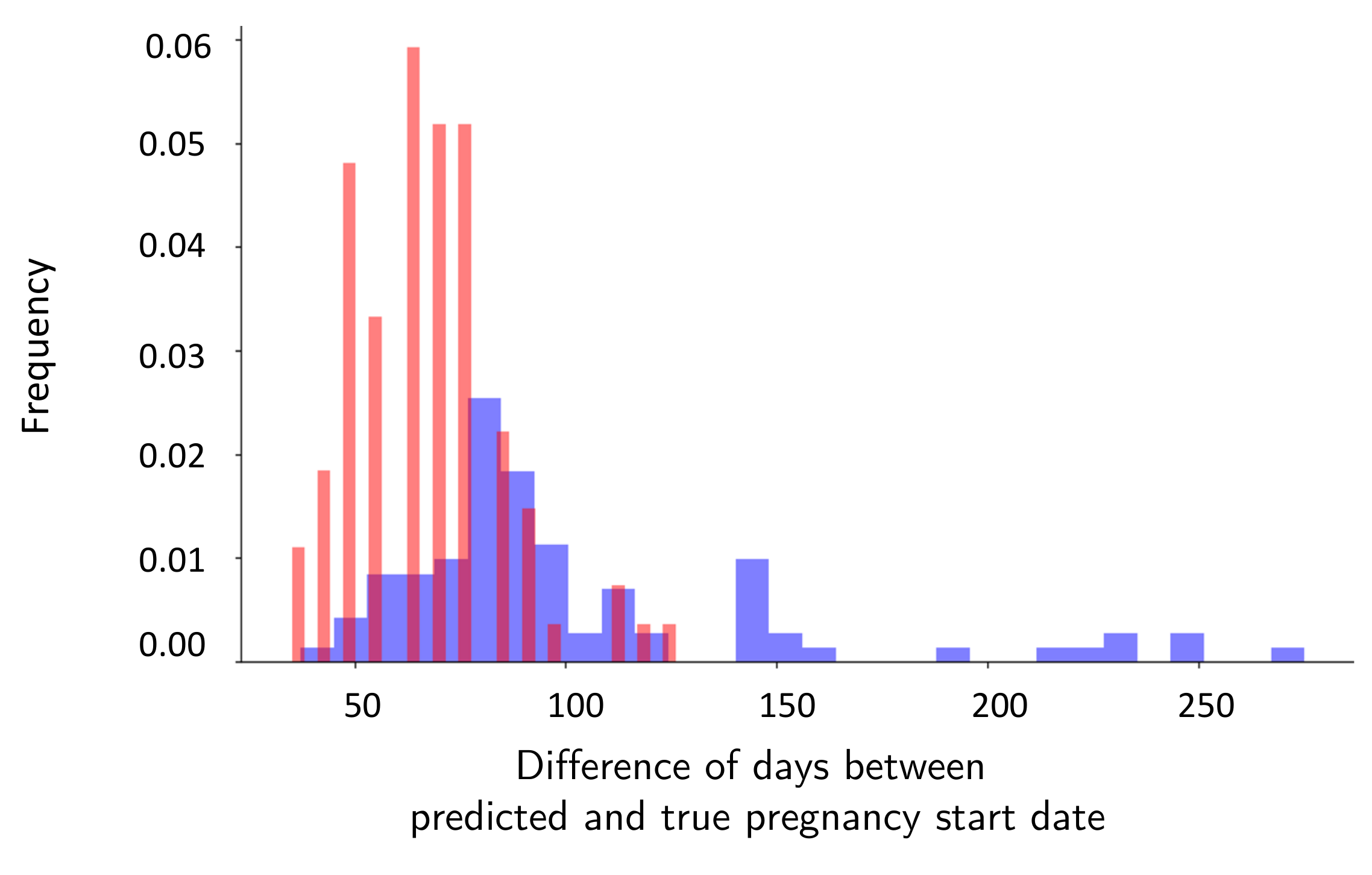}\label{fig:better_subset}}%
  }
\end{figure}

\subsection{Additional Results for Pregnancy Complication Prediction Retrospective Evaluation}

We include additional results for our retrospective evaluation of the pregnancy complications algorithm.

 In Table~\ref{tab:subgroup_metrics} we show the performance of our proposed predictor GROUP-Lasso that conditions on the patient's prior history of disease and predicts using separate Lasso models for each sub-group compared to the global Lasso model.
 Modeling outcomes for separate groups increases accuracy as the predictions become better calibrated but sacrifices ranking ability in terms of AUC. The advantage of GROUP-Lasso is that the features surfaced as explanations by the sub-group models show information beyond prior history may be useful for the care managers. Therefore, we use the global Lasso model to make predictions but use GROUP-Lasso to surface features as an explanation.

\begin{figure}[H]
    \centering
    \begin{tabular}{cc}
          & \includegraphics[width=\textwidth]{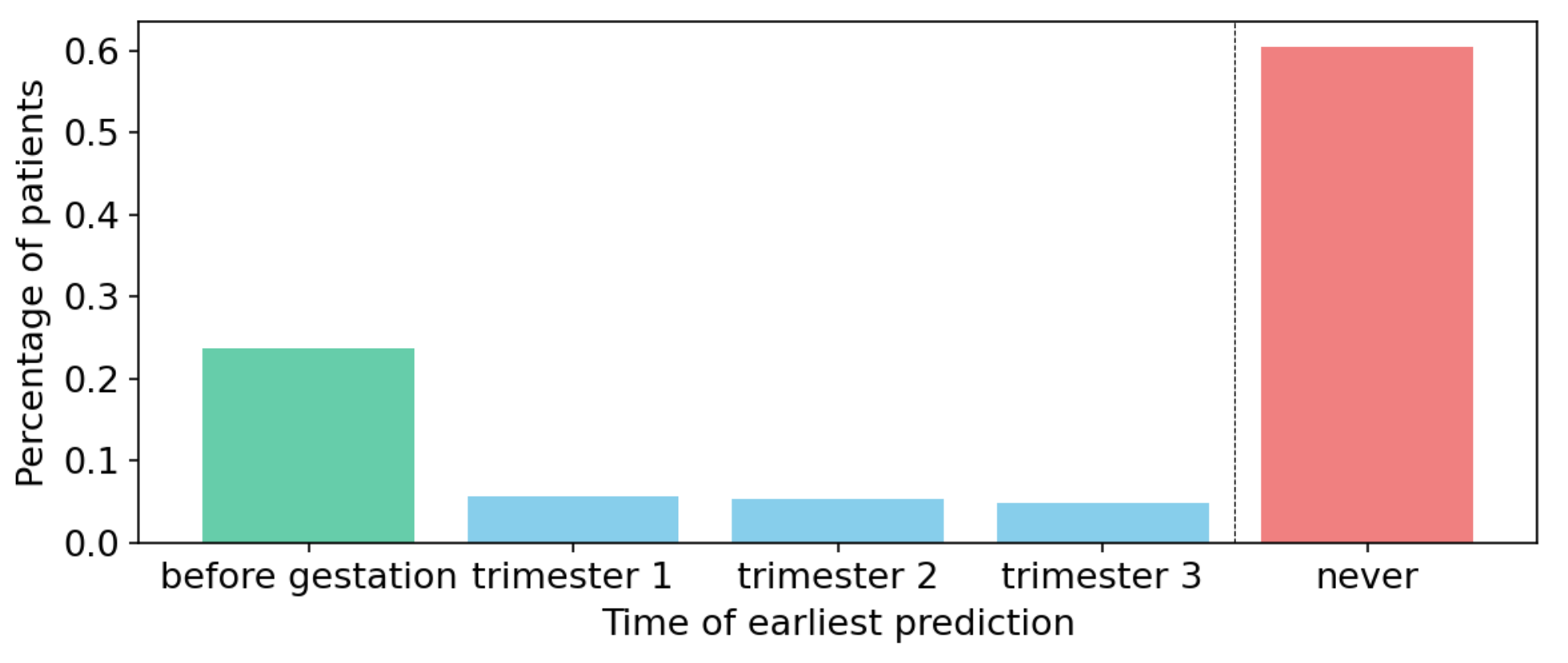} 
    \end{tabular}
    \caption{Distribution of earliest risk predictions of the Lasso model for members who have pregnancy complications of either gestational hypertension or diabetes. Members are placed into the respective bucket of the time the Lasso model first predicts a complication or if it never makes such a prediction. 
    } 
    \label{fig:how_early_risk}
\end{figure}

\begin{figure}[H]
    \centering
    \begin{tabular}{cc}
         (a) & \includegraphics[width=0.7\textwidth]{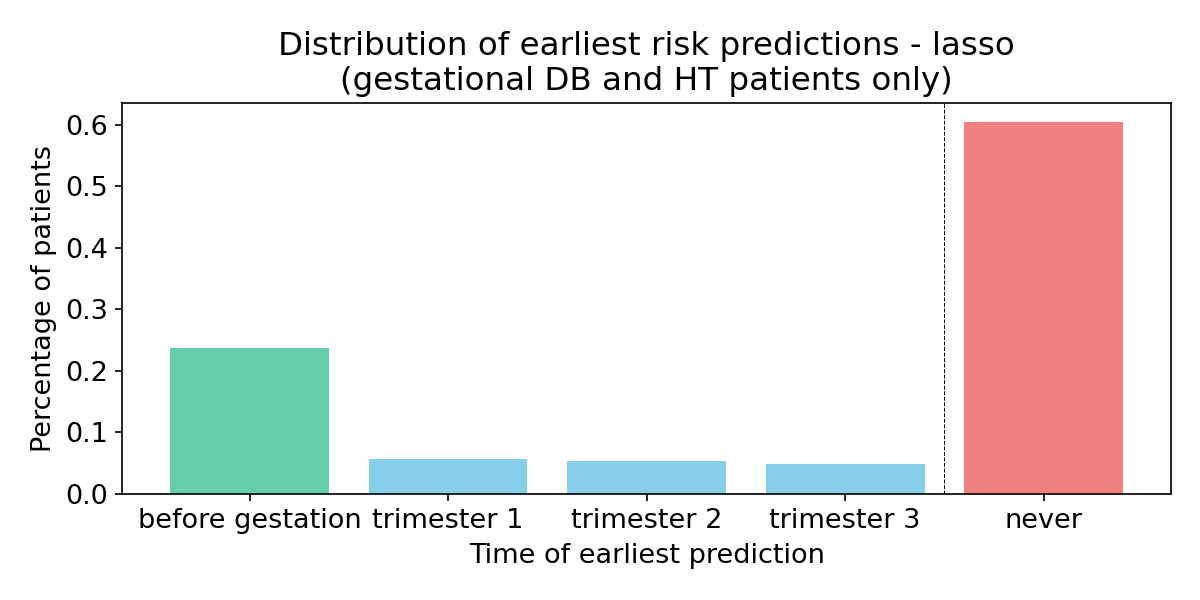} \\
         (b) & \includegraphics[width=0.7\textwidth]{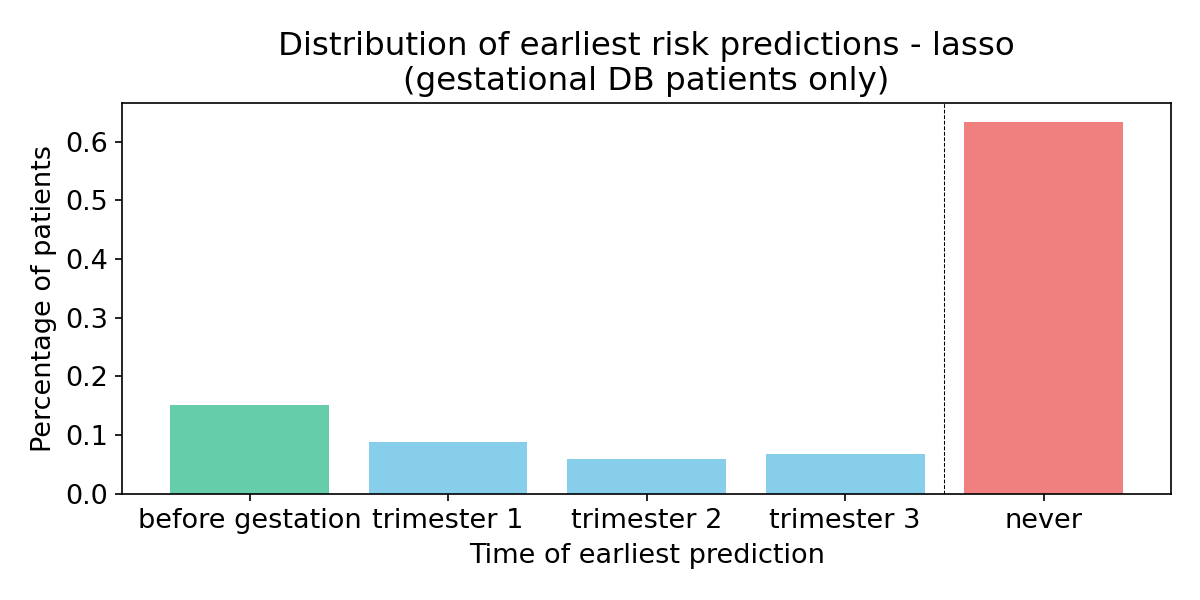} \\
         (c) & \includegraphics[width=0.7\textwidth]{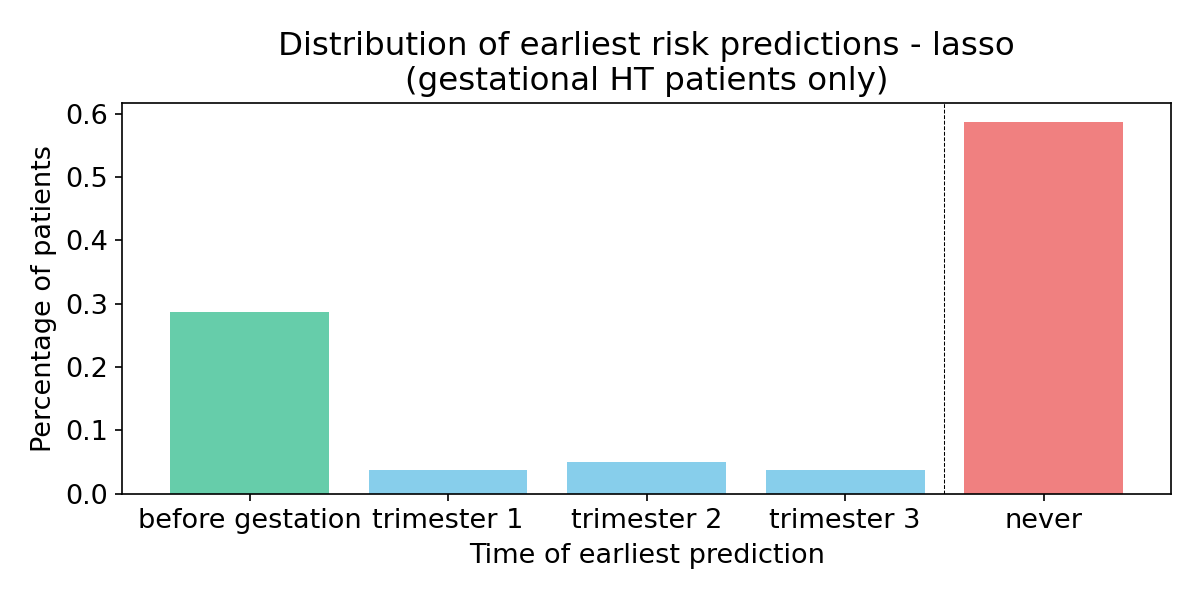}
    \end{tabular}
    \caption{Distribution of earliest risk predictions for members at risk of (a) both gestational DB and HT, (b) only gestational DB, and (c) only gestational HT.}
    \label{fig:how_early_risk_more}
\end{figure} 

\begin{table}[H]
    \centering
    \begin{tabular}{cccc}
        \hline
        &  & GROUP-Lasso & Lasso \\
        \hline
        \multirow{2}{*}{\textbf{History of DB}} & AUROC & 0.675 & 0.706 \\
        \multirow{2}{*}{} & Accuracy & 0.622 & 0.570 \\
        \hline
        \multirow{2}{*}{\textbf{History of HT}} & AUROC & 0.6573 & 0.708 \\
        \multirow{2}{*}{} & Accuracy & 0.708 & 0.647 \\
        \hline
        \multirow{2}{*}{\textbf{History of DB+HT}} & AUROC & 0.635 & 0.757 \\
        \multirow{2}{*}{} & Accuracy & 0.624 & 0.568 \\
        \hline
        \multirow{2}{*}{\textbf{No history of DB/HT}} & AUROC & 0.596 & 0.667 \\
        \multirow{2}{*}{} & Accuracy & 0.793 & 0.780 \\
        \hline
    \end{tabular}
    \caption{Evaluation metrics for each subgroup, comparing the subgroup model GROUP-Lasso to the global model Lasso. We partition the training and test sets based on member's prior history of gestational diabetes (DB) or gestational hypertension (HT). We then train a Lasso model (Lasso) on all the partitions and then evaluate it on each test partition. We then train 4 different Lasso models (GROUP-Lasso) on each partition in the training and data and then evaluate. }
    \label{tab:subgroup_metrics}
\end{table}

\subsection{Additional Details for User Studies}

We include additional details of our user studies.

\begin{figure}[h]
    \centering
    \begin{tabular}{cc}
        \includegraphics[width=0.45\textwidth]{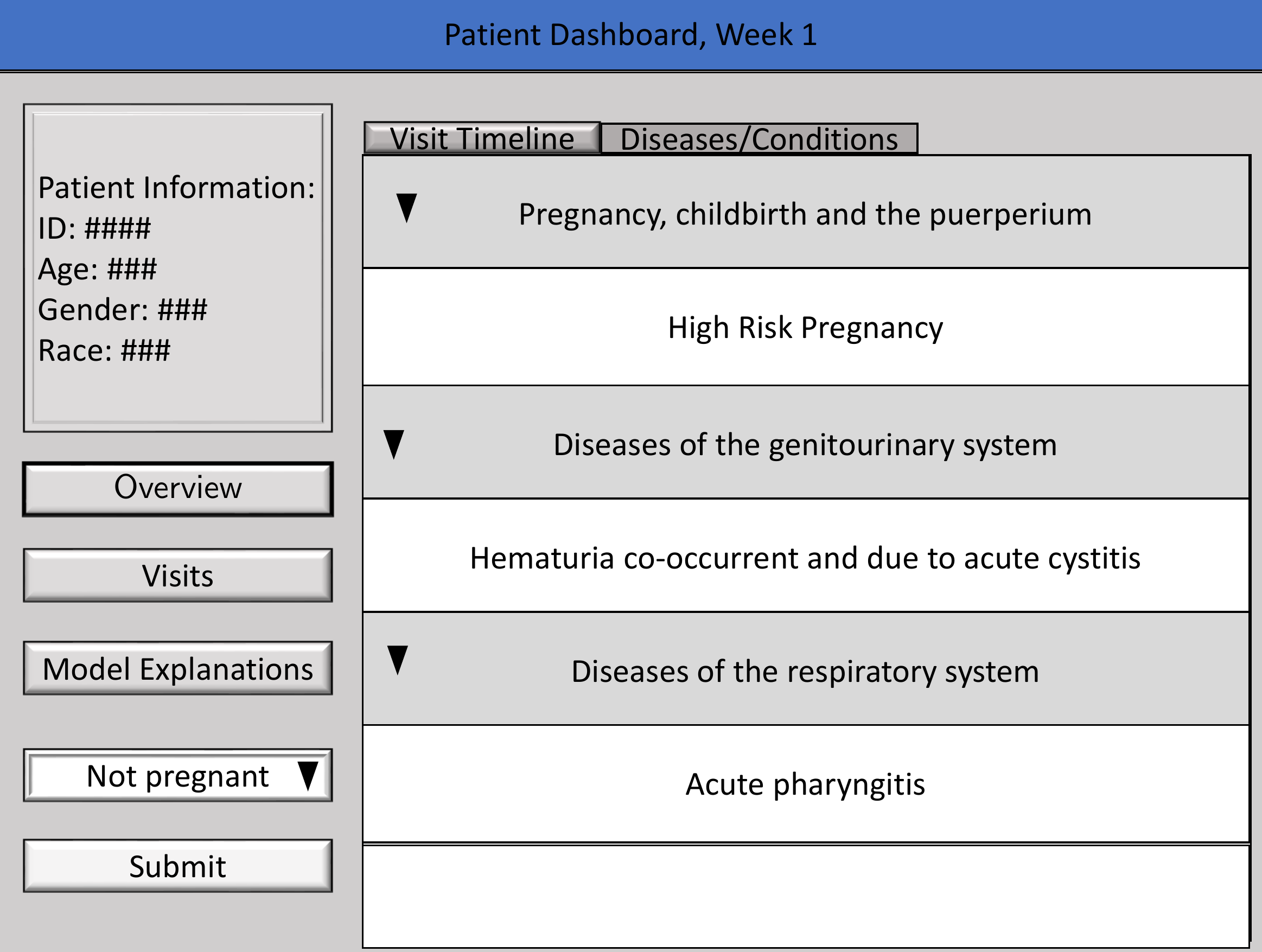} &    \includegraphics[width=0.45\textwidth]{figures/risk_interface_visits.pdf} \\
                (a) Pregnancy Identification Interface & (b) Pregnancy Complications Interface
    \end{tabular}
    \caption{Patient dashboard sketch for the user study on (a) pregnancy identification with \hapi and (b) pregnancy complications classification. In sub-figure (a) the user interface consists of a left panel containing demographic information and three views: Overview, Visits, and Model Explanation (evidence). We show the subtab Diseases/Conditions from the overview view where the nurse can find the ICD codes for each condition and disease. In sub-figure (b) we update the user interface taking into account feedback from the care managers by integrating the model predictions and evidence into the visit and overview views. On the left panel, patient information is shown as well as the model prediction and history of prior complications. We show in the figure the Diseases/Conditions view, we color ICD codes that are positively associated with complications with red (intensity varies with correlation) and those negatively associated with complications with green. }
    \label{fig:preg_dashboard}
\end{figure} 

\clearpage

\vspace{2cm}
\begin{table}[H]
	\centering
	\begin{tabular}{cc}
		(a) & \begin{tabular}{cp{0.35\textwidth}|p{0.35\textwidth}|}
			\hline
			\textbf{Trial} & \textbf{Members with information beyond prior knowledge} & \textbf{Examples} \\
			\hline
			A & 55.6\% (10 members) & oligohydramnios, premature delivery, blood clot, cervical issues, large baby, fetal hereditary disease, cancer, abnormal heart rate, first pregnancy \\
			&&\\
			B & 33.3\% (6 members)  & mental health / potential for postpartum depression, fetal abnormality, obesity, cervical issue, first pregnancy \\
			&&\\
			C & 66.7\% (12 members) & elevated glucose during current pregnancy / abnormal glucose code, h/o premature delivery, first pregnancy, twins, polycystic ovary syndrome, cervical incompetence (risk for preterm birth), elevated protein labs in current pregnancy \\
			\hline
		\end{tabular}\\
		\\
		(b) & \begin{tabular}{cp{0.35\textwidth}|p{0.35\textwidth}|}
			\hline
			\textbf{Trial} & \textbf{Members with information beyond prior knowledge} & \textbf{Examples} \\
			\hline
			A & 55.6\% (10 members) & previous retained placenta, home injections, pulmonary embolism, pre-term delivery, thalassemia, elevated glucose, asthma, hypothyroidism, infertility, uterine leimyoma, anemia, musculoskeletal disease, polycystic ovary syndrome, methadone \\
			&&\\
			B & 27.8\% (5 members)  & asthma, pre-term delivery, hypothyroidism, obesity, fibroids, infertility \\
			&&\\
			C & 61.1\% (11 members) & firbroids, previous losses, pre-term delivery, thyroid disease, cardiac murmur, Rhesus -, obesity, cardiac concern, hypothyroidism, infertility \\
			\hline
		\end{tabular}
	\end{tabular}
	\caption{Summary of members whose nurse notes contain information beyond prior knowledge (age, race, prior history of DB/HT) for nurse 1 (a) and nurse 2 (b). }
	\label{tab:risk_exp_info_beyond_prior}
\end{table}

\begin{table}[H]
	\centering
	\begin{tabular}{|p{0.2\textwidth}|p{0.2\textwidth}|p{0.2\textwidth}|p{0.2\textwidth}|}
		\hline
		\textbf{Category} & \textbf{Sub-category} & \textbf{\# of members} & \textbf{Simulation start date range} \\
		\hline
		Pregnant members detected by model & Detected early within reasonable time (at least 1 month after $t_{start}$) & 2 & $[t_{start}^{*}-3\text{ weeks}$, $t_{start}^{*}]$ \\
		&&&\\
		& Detected too early (before 1 month after $t_{start}$) & 2 & $[t_{start}^{*}-3\text{ weeks},$ $t_{start}^{*}]$ \\
		\hline
		Pregnant members detected by code & -- & 4 & $[t_{start}^{*}-1\text{ week},$ $t_{start}^{*}+2\text{ weeks}]$ \\
		\hline
		Non-pregnant members & Detected not pregnant & 3 & $\left[\tau^{0}, \tau^{'}-5\text{ weeks}\right]$ \\
		&&&\\
		& Detected pregnant & 1 & $[t_{start}^{*}-3\text{ weeks},$ $t_{start}^{*}]$ \\
		&&&\\
		\hline
	\end{tabular}
	\caption{Distribution of members and range of simulation start dates for pregnancy identification study. Note that for members detected pregnant, we sample within a fixed window around $t^{*}_{start}$, the pregnancy start time inferred by the algorithm; this ensures that the pregnancy prediction triggers during the 5-week simulation period. For members detected not pregnant, we sample from $\tau^{0}$, the time of the first sampled data point, to $\tau^{'}-5\text{ weeks}$, 5 weeks prior to the last sampled data point, so there is sufficient data for simulation.}
	\label{tab:preg_id_study_patients}
\end{table} 

\begin{table}[H]
	\centering
	\begin{tabular}{cccc}
		\hline
		\textbf{Outcome} & \shortstack{\textbf{Correct} \\ \textbf{Prediction?}} & \textbf{Prior History?} & \shortstack{\textbf{Number of}\\\textbf{Members}}\\
		\hline
		Gestational DB & Yes & No DB history & 3 \\
		Gestational HT & Yes & No HT history & 3 \\
		No complication & Yes & No DB or HT history & 3 \\
		Gestational DB & No & No DB history & 1 \\
		Gestational HT & No & No HT history & 1 \\
		No complication & No & No DB or HT history & 1 \\
		Gestational DB & Yes & DB history & 1 \\
		Gestational HT & Yes & HT history & 1 \\
		No complication & Yes & DB+HT history & 1 \\
		Gestational DB & No & DB history & 1 \\
		Gestational HT & No & HT history & 1 \\
		No complication & No & DB+HT history & 1 \\
		\hline
	\end{tabular}
	\caption{Distribution of members for pregnancy risk factor study. We sample members evenly across the three outcomes, with more members without prior history to reflect the overall distribution. We include both correct and incorrect predictions to evaluate how well nurses filter errors.}
	\label{tab:risk_study_pat_distribution}
\end{table} 

\end{document}